\documentclass{article}

\usepackage{amsmath,amsfonts,bm}

\def\eqref#1{equation~\ref{#1}}

\def\1{\bm{1}}

\def\rz{{\textnormal{z}}}

\def\vx{{\bm{x}}}

\DeclareMathAlphabet{\mathsfit}{\encodingdefault}{\sfdefault}{m}{sl}
\SetMathAlphabet{\mathsfit}{bold}{\encodingdefault}{\sfdefault}{bx}{n}

\def\gO{{\mathcal{O}}}

\def\sD{{\mathbb{D}}}

\def\sR{{\mathbb{R}}}

\def\wh#1{\widehat{#1}}

\usepackage{subcaption}
\usepackage[boxed]{algorithm2e}
\usepackage{forest}
\usepackage{footnote}

\newcommand\given[1][]{\:#1\vert\:}

\usepackage{microtype}
\usepackage{graphicx}
\usepackage{booktabs} 

\usepackage[accepted]{icml2019}

\icmltitlerunning{Hierarchical Routing Mixture of Experts}
\begin{document}

\twocolumn[
\icmltitle{Hierarchical Routing Mixture of Experts}



\icmlsetsymbol{equal}{*}

\begin{icmlauthorlist}
\icmlauthor{Wenbo Zhao}{ece}
\icmlauthor{Yang Gao}{ece}
\icmlauthor{Shahan Ali Memon}{cs}
\icmlauthor{Bhiksha Raj}{cs}
\icmlauthor{Rita Singh}{cs}
\end{icmlauthorlist}

\icmlaffiliation{ece}{Department of Electrical and Computer Engineering, Carnegie Mellon University, Pittsburgh, PA, USA}
\icmlaffiliation{cs}{School of Computer Science, Carnegie Mellon University, Pittsburgh, PA, USA}

\icmlcorrespondingauthor{Wenbo Zhao}{wzhao1@andrew.cmu.edu}

\icmlkeywords{Regression, Hierarchical Mixture Models, Algorithm}

\vskip 0.3in
]



\printAffiliationsAndNotice{}  

\begin{figure*}[!h]
\centering
\begin{subfigure}[t]{0.3\linewidth}
    \includegraphics[width=\linewidth]{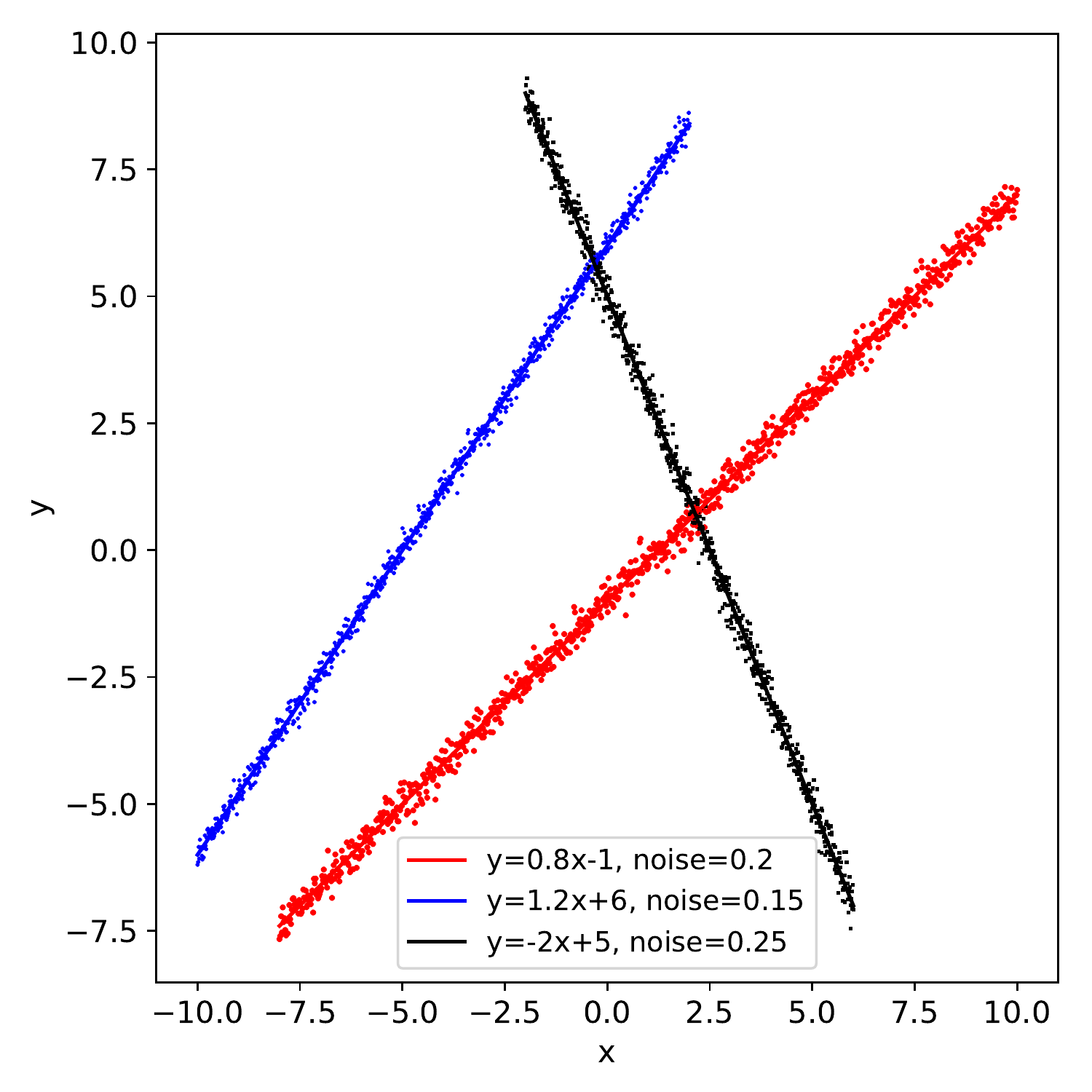}
    \caption{}
    \label{sfig:data_3lines}
\end{subfigure} 
\begin{subfigure}[t]{0.3\linewidth}
    \includegraphics[width=\linewidth]{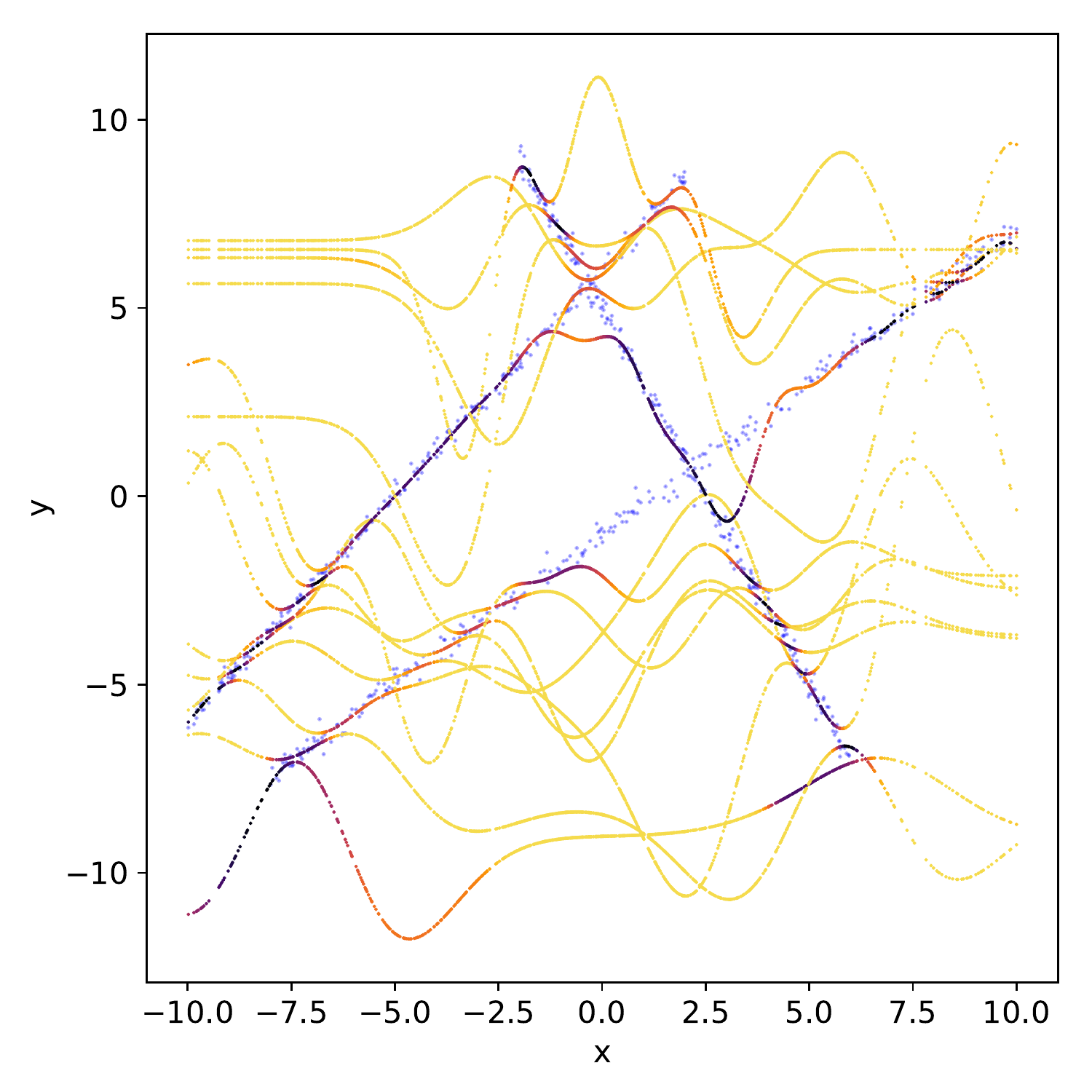}
    \caption{}
    \label{sfig:3lines_plots_expert}
\end{subfigure}  
\begin{subfigure}[t]{0.3\linewidth}
    \includegraphics[width=\linewidth]{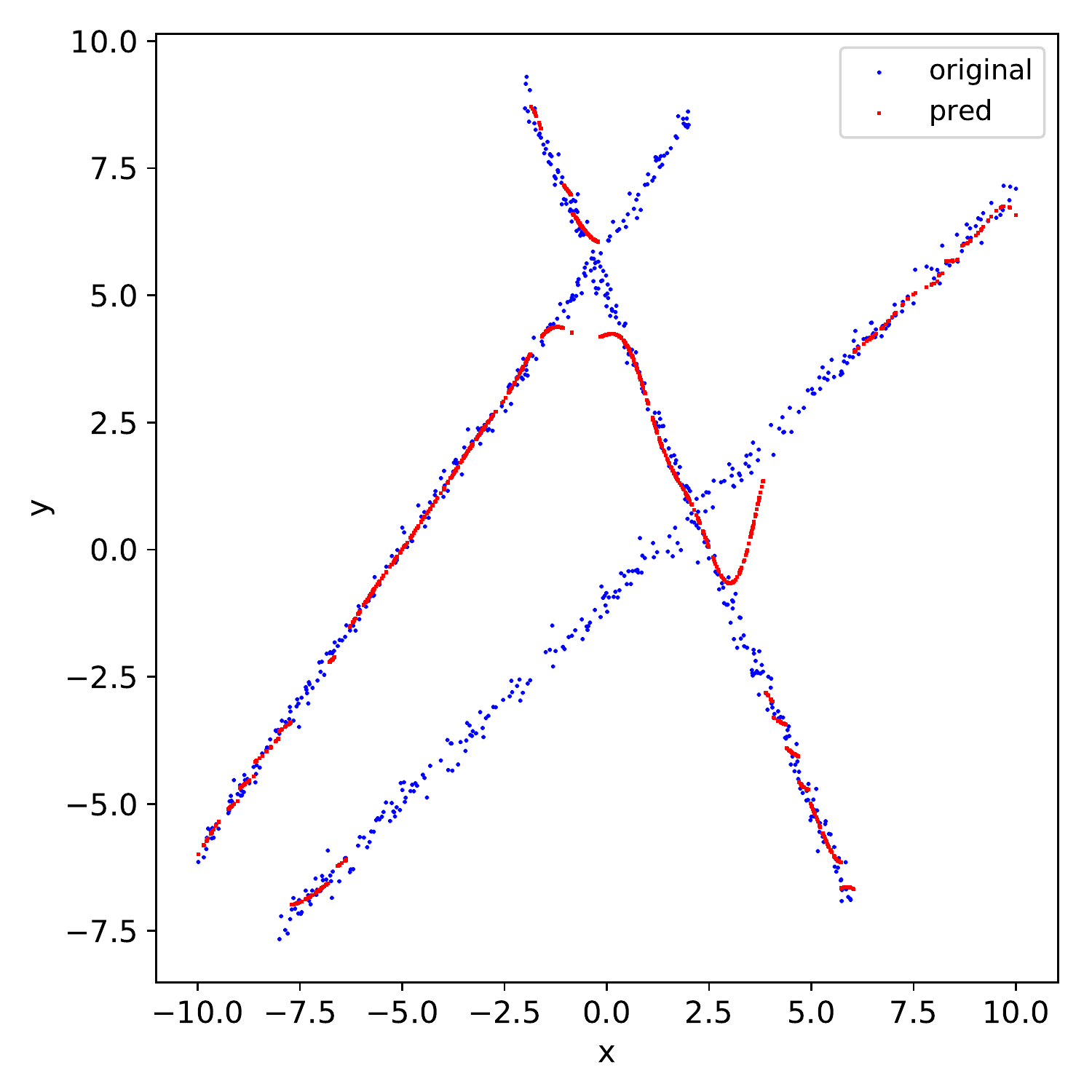}
    \caption{}
    \label{sfig:3lines_plots_expert_top}
\end{subfigure}
\caption{(a) A toy example: synthetic 3-lines data with different amount of noises.
(b) Predictions made by experts in our HRME model. Each curve represents the prediction made by one expert. Darker color indicates stronger prediction confidence.
(c) Prediction made by our HRME model via selecting the top-$1$ experts.}
\label{fig:3lines_plot_collection}
\end{figure*}

\begin{abstract}
In regression tasks the distribution of the data is often too complex to be fitted by a single model. In contrast, partition-based models are developed where data is divided and fitted by local models. These models partition the input space and do not leverage the input-output dependency of multimodal-distributed data, and strong local models are needed to make good predictions. Addressing these problems, we propose a binary tree-structured hierarchical routing mixture of experts (HRME) model that has classifiers as non-leaf node experts and simple regression models as leaf node experts. The classifier nodes jointly soft-partition the input-output space based on the natural separateness of multimodal data. This enables simple leaf experts to be effective for prediction. Further, we develop a probabilistic framework for the HRME model, and propose a recursive Expectation-Maximization (EM) based algorithm to learn both the tree structure and the expert models. Experiments on a collection of regression tasks validate the effectiveness of our method compared to a variety of other regression models.
\end{abstract}

\section{Introduction}



One of the challenges in modeling a regression task is that of dealing with data with complex distributions.  The distribution can be multi-modal, rendering any single regression model highly biased.
For instance, Figure~\ref{sfig:data_3lines} shows a synthetic data set uniformly sampled from three intersecting lines with different amount of noise.
A single regression model would fail to capture the multi-modality of this data and yield poor performance.
This necessitates another strategy, of divide and conquer, to partition the input space into simple sub-regions and assign a regression model to each sub-region.

Many models take this strategy. For example decision trees~\cite{loh2014fifty} and random forests~\cite{breiman2001random} divide the input space by hard-partition of feature dimensions, and make piece-wise linear predictions on each partition. Mixture models~\cite{bailey1994fitting} and mixtures-of-experts~\cite{jacobs1991adaptive} perform soft-partition on the input space and assign regression models to each of the partitions. In particular, the mixture of experts models are tree-structured models with a gating mechanism to partition the input space and a collection of experts at the leaves to make local predictions.

Although well-studied and have been proven effective,  these models do not leverage the input-output dependency of the data distributions. For instance, as in our toy example, different regions of the output space (the $y$ label) correspond to different modes of the data.
Solely partitioning the input space would fit multiple modes of the data into each partition, still requiring complex regression models to capture the input-output relation in each of them. This problem can be avoided by jointly partitioning both the input and output spaces, such that each partition only requires a simpler local regression.  This is the motivation behind our work.

Addressing the above-mentioned issues of conventional partition-based regression methods, we propose a {\em hierarchical routing mixture of experts} (HRME) model, which separates output variables modes by jointly partitioning the input and  output spaces, and makes probabilistic inferences by assigning simple regression models to each of the resultant partitions.
Our HRME model can be viewed as a new member of the family of hierarchical mixture of experts (HME)~\cite{jordan1994hierarchical} models. It is binary-tree structured, and has two types of experts---the non-leaf node experts and leaf node experts.
The {\em non-leaf node experts} function as a new gating mechanism to soft-partition the data based on their modes, defined on the {\em joint} distribution of input and output variables. The partitioning is performed by node-specific binary classifier. Together, the classifiers in the non-leaf nodes hierarchically partition the space into number of regions, each of which corresponds to a leaf in the tree, and within which the relationship between input and output variables is ideally unimodal. The {\em leaf node experts} make predictions on each resulting partition. If the data is well partitioned, these leaf node experts can now be relatively simple.

However, the actual distribution of the data and its modes are unknown {\em a priori}. Consequently, the binary classes for each classifier (non-leaf) node are unknown. This effectively makes the partition of the output space itself a variable to be determined.
To address this, we develop a probabilistic framework for our HRME model, and propose a recursive Expectation-Maximization (EM) based algorithm to optimize the joint input-output partition, the various expert models, as well as the tree structure.
To the best of our knowledge, this new joint-partition based gating mechanism for HME models has not been studied yet. The closest relevant literature is by~\citet{memon2018neural} which partitions the space solely based on the {\em output} value to determine its optimal discretization. Our HRME model, on the other hand, uses a joint partition to determine the optimal data allocation to the leaf experts, and our model is globally optimized rather than locally optimized.

We test our model on a collection of standard regression tasks, and the results validate the effectiveness of our model compared to HME and other regression models.
Our contributions are: (1) we propose a new gating mechanism via joint-partition of both input space and output space to separate the modes of complexly distributed data, making simple regression models effective for predictions;
(2) we develop a recursive EM algorithm to jointly optimize the partition, the expert models, as well as the tree structure.

\section{Related Work}
Decision trees~\cite{loh2014fifty,breiman2017classification,quinlan1986induction} are a family of supervised learning methods that utilize a partition on the input feature space and make piece-wise linear predictions. Based on them, random forests~\cite{breiman2001random,liaw2002classification} take an ensemble learning approach by aggregating a collection of decision trees to reduce the over-fitting tendency of a single decision tree. A pertaining issue with these tree-based methods is that they rely on hard partitions and piece-wise linear predictions, which can lead to discontinuities and high biases in predictions.

On the other hand, the mixture of experts (ME) models are a family of probabilistic tree-structured models with a gating mechanism and a collection of experts at the leaves. The gating mechanism is responsible for soft partitioning the input space into sub-regions such that a local expert models the distribution of each sub-region~\cite{yuksel2012twenty}. 
The flexibility of the ME family embraces a rich variety of gating mechanisms and expert models. Examples include hierarchical mixture of experts (HME)~\cite{jordan1994hierarchical} which employs a binary tree structure, Bayesian HME~\cite{bishop2002bayesian} with a Bayesian treatment, mixture of Gaussian processes (HME-GP)~\cite{tresp2001mixtures,rasmussen2002infinite,yuan2009variational,nguyen2014fast}, mixture of support vector machines (HME-SVM)~\cite{lima2007hybridizing,cao2003support}, to name only a few.

The ME models have three issues: (1) the gating mechanism does not explicitly leverage the input-output dependencies of the data. Rather, it performs probabilistic input-space partitioning, based on assumed data distributions such as the multinomial distribution~\cite{jordan1994hierarchical}, Gaussian distribution~\cite{yuan2009variational}, Dirichlet process~\cite{rasmussen2002infinite}, Gaussian process~\cite{tresp2001mixtures}, etc; (2) in ME models strong experts are often needed to gain good performance~\cite{yuksel2012twenty}; (3) the structure of the ME models, namely the tree depth and the number of experts, is often optimized through extra procedures, such as pruning~\cite{waterhouse1995pruning} and Bayesian model selection~\cite{bishop2002bayesian,kanaujia2006learning}. This increases the complexity of model learning.

We address the issues with these conventional methods by (1) joint soft-partition of the input-output space based on the natural separability of the multi-modal data and (2) joint optimization on the tree structure and the expert models without extra pruning procedures.

\section{Hierarchical Routing Mixture of Experts}
In this section, we present the specifications of the HRME model, formulate the optimization objective, and develop the optimization algorithm.

\subsection{Model Specification}
\label{ssec:tree_model_spec}

\begin{figure}[t]
\centering
\scalebox{0.8}{
\includegraphics[width=0.8\columnwidth]{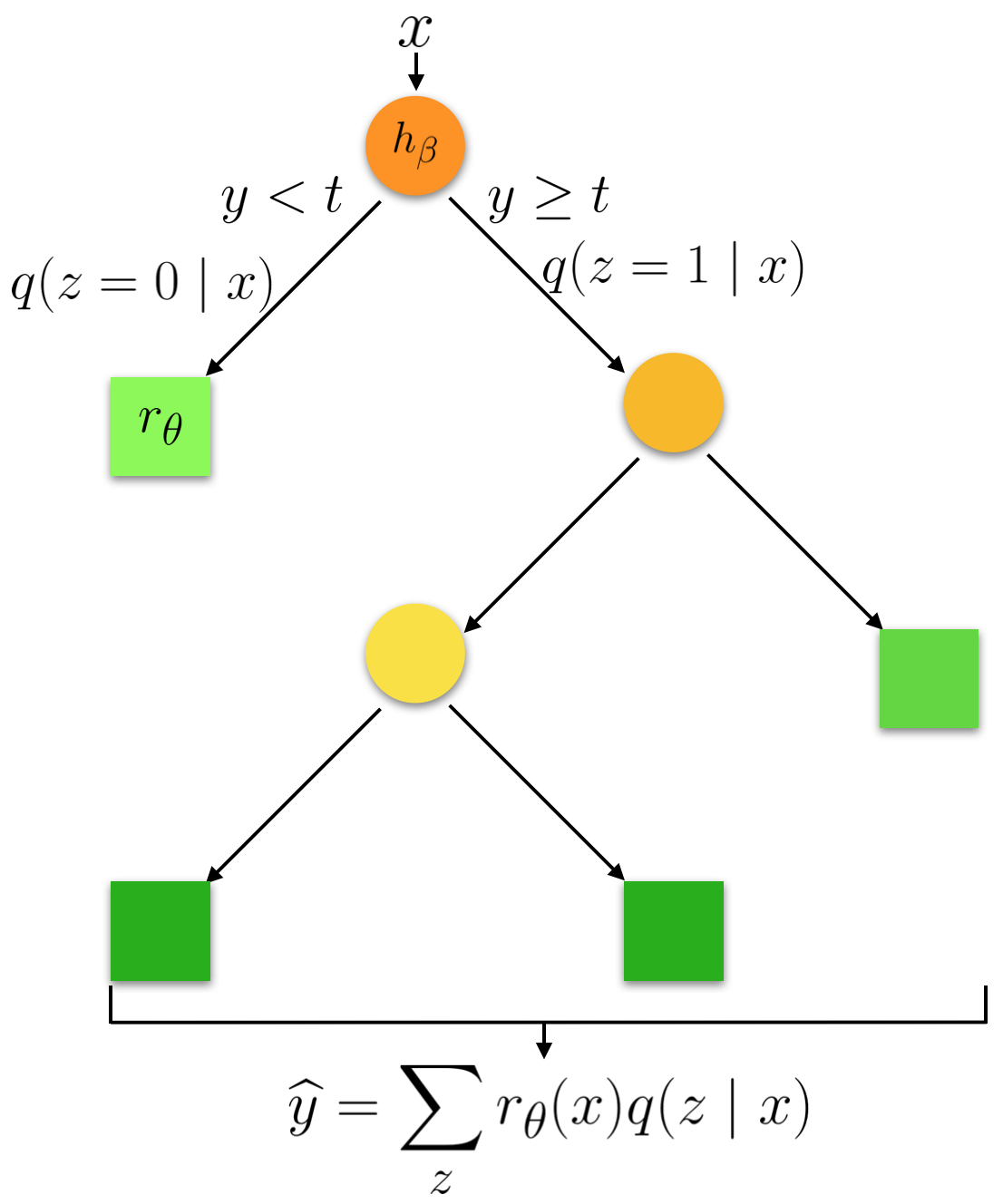}
}
\caption{Illustration of the HRME model. It is a probabilistic binary tree. Each non-leaf node (circle) carries a classifier $h_{\beta}$ and a partition threshold $t$, and each leaf node (square) carries a regressor $r_{\theta}$. Prediction is made via probabilistic combination of leaf regressors. Model is learned via recursive EM.}
\label{fig:hrme_model}
\end{figure}

Figure~\ref{fig:hrme_model} shows the structure of the tree model. It is a binary tree.
Each non-leaf node is equipped with a classification expert, which is a binary classifier in this case.
Each leaf node is equipped with a regression expert, which is a simple linear model.
The basic assumption here is that the complexly distributed multi-modal data is nevertheless locally (and non-linearly) separable, and hence the non-leaf experts of the tree function as a routing mechanism to partition the data into subsets of simple (uni-modal) distributions, and route each subset to a simple leaf expert to make predictions.

We denote the input features as $\vx \in \sR^d$ and the continuous output label as $y \in \sR$.
In order to route the data in such fashion, it requires to determine the optimal classifier at each non-leaf node. However, we do not have the data class information beforehand, i.e., we do not know how data can be locally separated.
As a remedy, we adopt a thresholding strategy---setting a threshold $t$ on $y$ such that $y=0$ if $y < t$ and $y=1$ otherwise. As a result, we assign binary classes to data via thresholding on $y$.
However, note that by doing so we effectively make $t$ a variable to be optimized, namely, we are not only partitioning on $\vx$, but also partitioning on $y$. We will explain optimization of this joint-partition in the later section.

At this point, let's assume we have known the optimal tree settings---that is, we know the tree structure (the depth and the number of nodes), and for each non-leaf node, the optimal splitting threshold $t^*$ and the classifier $h_{\beta^*}$ parameterized by the optimal parameter $\beta^*$, and for each leaf node, the regressor $r_{\theta^*}$ parameterized by the optimal parameter $\theta^*$. We then explain the prediction of $y$ given an input $\vx$.

Specifically, for notation convenience, we assume the nodes are numbered such that for any two nodes $n_i$ and $n_j$, if $i < j$, $n_i$ occurs either to the left of $n_j$ or above it in the tree.
Each node $n_i$ carries a classifier $h_{\beta_{n_i}^*}: \vx \mapsto \{n_{i+1}, n_{i+2}\}$, which assigns any instance with input $\vx$ to one of the children nodes $n_{i+1}$ or $n_{i+2}$.
We introduce a binary-valued random variable $\rz_{n_i} \in \{0, 1\}$ to indicate $\vx$ being assigned to $n_i$ or not. Then, the corresponding likelihood of $\vx$ being assigned to node $n_i$ is estimated by the classifier on node $n_{i-1}$
\begin{align}
q(\rz_{n_i} \given \vx) \equiv q(\rz_{n_i}=1 \given \vx) \longleftarrow h_{\beta_{n_{i-1}}^*}(\vx).
\label{eq:qn_x}
\end{align}
Next, we would like to know the likelihood of a data point $\vx$ being routed to a specific leaf.
Denote the chain from root $l_1 \equiv n_0$ to leaf $l_k$ as $l_1 \rightarrow \ldots \rightarrow l_k$, then the likelihood of $\vx$ being assigned to leaf $l_k$ is
\begin{align}
    q(\rz_{l_k} \given \vx) = \sum_{\rz_{l_1}}\ldots\sum_{\rz_{l_{k - 1}}} q(\rz_{l_1}, \ldots, \rz_{l_k} \given \vx).
    \label{eq:qz_x_sum_prod}
\end{align}
Applying the sum-product rule and using the conditional dependency to (\ref{eq:qz_x_sum_prod}) yield
\begin{align}
    q(\rz_{l_k} \given \vx) = \prod_{j=1}^{k-1}q(\rz_{l_{j+1}} \given \rz_{l_{j}}, \vx).
    \label{eq:qz_x}
\end{align}
For leaf $l_k$, it carries a regressor $r_{\theta_{l_k}^*}$ such that the prediction $\wh{y}_{l_k} = r_{\theta_{l_k}^*}(\vx)$. Then, an estimate of $y$ is given by the expectation of the predictions over all leaves
\begin{align}
    \wh{y} = \sum_{l_k \in \textnormal{leaves}} r_{\theta_{l_k}^*}(\vx) q(\rz_{l_k} \given \vx),
    \label{eq:y_hat}
\end{align}
and the corresponding conditional density for leaf $l_k$ is
\begin{align}
    p(y \given \rz_{l_k}, \vx) \leftarrow r_{\theta_{l_k}^*}(\vx).
    \label{eq:py_zx}
\end{align}

\subsection{Learning Algorithm}
\label{ssec:learn_tree}
From the previous section, we have shown that in order to make predictions using the tree, we need to determine the optimal tree settings, i.e., the tree structure $\{n_i\}$, the non-leaf node thresholds $\{t_{n_i}\}$, the classifier parameters $\{\beta_{n_i}\}$, and the leaf node regressor parameters $\{\theta_{n_i}\}$.

We adopt a maximum-likelihood approach. Specifically, our objective is to maximize the log-likelihood for each $\vx$
\begin{align}
    \max &\log p(y \given \vx) \label{eq:obj}\\
    = &\log p(y \given \vx) \frac{q(\rz \given \vx)}{q(\rz \given \vx)} \sum_{\rz} q(\rz \given \vx)\nonumber\\
    = &\sum_{\rz} q(\rz \given \vx) \log \frac{p(y, \rz \given \vx)}{p(\rz \given y, \vx)} \frac{q(\rz \given \vx)}{q(\rz \given \vx)} \nonumber\\
    = &\sum_{\rz} q(\rz \given \vx) \log \frac{p(y, \rz \given \vx)}{q(\rz \given \vx)} + \label{eq:ELBO}\\
      &\sum_{\rz} q(\rz \given \vx) \log \frac{q(\rz \given \vx)}{p(\rz \given y, \vx)}, \label{eq:KL}
\end{align}
where $q(\rz \given \vx)$ is an estimate for the true assignment mass $p(\rz \given \vx)$; (\ref{eq:ELBO}) is commonly referred to as the evidence lower bound (ELBO) which need to be improved to maximize the log-likelihood (\ref{eq:obj}); (\ref{eq:KL}) is the Kullback-Leibler divergence which measures the distance of two probability masses and is always greater than or equal to zero.

Therefore, it is natural to apply the expectation-maximization (EM) method to optimize (\ref{eq:obj}).
Specifically, in the E-step, we compute the ELBO (\ref{eq:ELBO}) for all the training instances
\begin{align}
    Q(p, q) &= \sum_{\vx} \sum_{\rz} q(\rz \given \vx) \log \frac{p(y, \rz \given \vx)}{q(\rz \given \vx)}\nonumber\\
    &= \sum_{\vx} \sum_{\rz} q(\rz \given \vx) \log \frac{p(y \given \rz, \vx) p(\rz \given \vx)}{q(\rz \given \vx)},
\end{align}
where $q(\rz \given \vx)$ is given by (\ref{eq:qz_x}), and $p(y \given \rz, \vx)$ is given by (\ref{eq:py_zx}) (for example, the leaf node gives a Gaussian distribution over $y$ if we assume a linear model with Gaussian noise). The true leaf node assignment mass $p(\rz \given \vx)$ is yet unknown. However, we can estimate it using the empirical frequency of the number of samples at the leaf node over the total number of training samples. This is a crude estimation, but we will provide a better strategy in the later part of this section.

In the M-step, we optimize the parameters to increase the ELBO (\ref{eq:ELBO}). Specifically, we optimize the non-leaf node expert to maximize the classification accuracy, and optimize the leaf-node expert to minimize the regression error.

However, as we mentioned in Section~\ref{ssec:tree_model_spec}, the data classes are not available, and the non-leaf node threshold $t$ is unknown. We provide an alternative approach to mitigate this difficulty. For each non-leaf node, we perform grid-search over the possible values of $t$, and for each $t$, we perform the M-step. The best $t$ value is then obtained as the one with maximum $Q$-value. Although different sampling strategies can be used when searching for $t$, in practice we find grid-search works well.

As we mentioned earlier, it is difficult to estimate the true leaf node assignment mass $p(\rz \given \vx)$. Although variational approximation may be used, we propose an empirically simpler strategy. Instead of using the $Q$-value as a global indicator of the optimality of the tree, we propose to use the negative mean-square-error
\begin{align}
    Q_{\textnormal{alternative}} = -\textnormal{mean} (y - \wh{y})^2,
    \label{eq:neg_mse}
\end{align}
where $\wh{y}$ is given by (\ref{eq:y_hat}).

\SetKwInOut{Parameter}{Parameter}
\SetKwProg{Fn}{Function}{}{}
\SetKwFunction{GrowTree}{GrowTree}
\SetKwFunction{GrowSubtree}{GrowSubtree}
\SetKwFunction{SplitData}{SplitData}
\SetKwFunction{TrainClassifier}{TrainClassifier}
\SetKwFunction{PredictProb}{PredictProb}
\SetKwFunction{TrainLeaf}{TrainLeaf}
\SetKwFunction{ComputeQ}{ComputeQ}

\begin{algorithm}[!t]
\caption{{\bf Recursive EM Learning of HRME}}
\label{alg:hrme}
    \KwIn{[\textit{data}], [\textit{root}]}
    \Parameter{$\{t\}$, classifier parameters, regressor parameters}
    \KwOut{HRME Tree}
	
    \Fn{\GrowTree{\textit{data\_list}, \textit{nodes\_per\_level}}}{
        
        \For{\textit{node} \textnormal{in} \textit{nodes\_per\_level}}{
            $\sD$ $\leftarrow$ \textit{data\_list}

            \textit{node\_l}, \textit{node\_r} $\leftarrow$ \GrowSubtree{\textit{node}}
            
            \For{$t$}{
                $\sD_l$, $\sD_r$ $\leftarrow$ \SplitData{$\sD$, $t$}
                
                \lIf{$\frac{\min(|\sD_l|, |\sD_r|)}{\#\text{ of total samples}} <\text{min\_leaf\_sample\_ratio}$}{continue}
                
                \textit{node}.\TrainClassifier{$\sD$, $t$}
                
                Propagate conditionals using Equation~(\ref{eq:qz_x})
                
                \textit{node\_l}.\TrainLeaf{$\sD_l$}
                
                \textit{node\_r}.\TrainLeaf{$\sD_r$}
                
                $Q$ $\leftarrow$ \ComputeQ \textnormal{ using Equation}~(\ref{eq:neg_mse})
            
            }
            
            \uIf{$Q > Q^*$}{
                $Q^* \leftarrow Q$
                
                \textit{data\_list} $\leftarrow$ \textnormal{[}$ \sD_l$, $\sD_r$\textnormal{]}
                
                \textit{nodes\_per\_level} $\leftarrow$ \textnormal{[}\textit{node\_l}, \textit{node\_r}\textnormal{]}
                
                \GrowTree{\textit{data\_list}, \textit{nodes\_per\_level}}
            }
            \Else{
                Delete the subtree
                
                continue
            }
        }
 	}
\end{algorithm}

The recursive EM algorithm is summarized in Algorithm~\ref{alg:hrme}.
We start from the root node, and grow the tree recursively in a \emph{depth-first} manner, i.e., from top to bottom, from left to right. Each time we grow a three-node subtree. We keep increasing the number of nodes until the lower bound $Q$ stops increasing or the ratio of the number of samples at the leaf to the total number of samples is below some preset threshold. 

\section{Experiments}
In this section, we evaluate our HRME model and the recursive EM algorithm on a collection of standard regression datasets. We describe the experiment settings and present the results for our method and a wide range of baseline methods.

\begin{table}[!h]
\caption{Dataset Statistics}
\label{tab:data_stat}
\vskip 0.5 \baselineskip
\centering
\begin{small}
\begin{sc}
\begin{tabular}{lcrr}
\toprule
Dataset & Feature Dim & Train & Test \\
\midrule
3-lines    & 1 & 1750 & 750 \\
Housing    &13 & 354  & 152 \\
Concrete   & 8 & 721  & 309 \\
CCPP       & 4 & 6697 & 2871 \\
Energy     & 28& 14803& 4932\\
Kin40k     & 8 & 10000& 30000\\
\bottomrule
\end{tabular}
\end{sc}
\end{small}
\end{table}

\subsection{Data}
For demonstration purpose, we synthesize a 3-lines dataset (as shown in Figure~\ref{sfig:data_3lines}).
For further evaluation, we select five other standard datasets that are commonly used in regression tasks.
Four of these datasets are from the UCI machine learning repository~\cite{Dua:2017}: the \texttt{CCPP} dataset~\cite{tufekci2014prediction,kaya2012local}, the \texttt{concrete} dataset~\cite{yeh1998modeling}, the \texttt{Boston housing} dataset~\cite{belsley2005regression} and the \texttt{energy} dataset ~\cite{candanedo2017data}, and one \texttt{kin40k} dataset~\cite{seeger2003fast,deisenroth2015distributed}.
The datasets range from small-sized to large-sized and from low-dimensional to high-dimensional. The statistics are shown in Table~\ref{tab:data_stat}. The division of train and test sets are either using the default split or using $0.7 : 0.3$ split.

\subsection{Models}
{\bf Baselines}
To promote a fair evaluation, we compare our method with a wide range of baselines: linear regression (LR), support vector regression (SVR), decision trees (DT), random forests (RF), hierarchical mixture of experts (HME) with strong Gaussian or Gaussian process experts, and multilayer perceptron (MLP). 
Each model carries a set of parameters to be estimated as well as hyperparameters (e.g., margin and kernels in SVR, depth and number of nodes in DT and RF, number of neurons and learning rate in MLP, etc.) to be tuned. We train the models on training sets, and fine-tune the hyperparameters using grid-search and three-fold cross validation on the training sets to obtain the best performance.
The models are implemented with scikit-learn toolkit~\cite{sklearn_api} or PyTorch~\cite{paszke2017automatic}.
For HME models, we obtain the best available results from the literature under the same experiment settings.

{\bf HRME}
For our HRME model, we train it following Algorithm~\ref{alg:hrme}. In our instantiation of the model, the non-leaf experts are support vector machines with radial basis function kernels (SVM-RBF). We choose two simple models for leaf experts, the linear regression model (referred to as HRME-LR) or the support vector regression model with radial basis function kernel (referred to as HRME-SVR).
Similar to the training of baselines, our models are also trained and fine-tuned on the same training sets following the same strategy with the baseline methods.
In addition, all non-leaf experts on the tree share the same hyperparameters, so are the leaf experts. Although it would be desirable to use different hyperparameters for nodes on different depth of the tree as the data size shrinks with the tree depth, in practice we find our model is robust to such variations.

\subsection{Results}

We evaluate our methods and the baseline methods with two metrics: the mean absolute error (MAE) and the root mean squared error (RMSE).

\begin{figure}[!t]
\centering
\begin{subfigure}[b]{0.3\linewidth}
    \includegraphics[width=\linewidth]{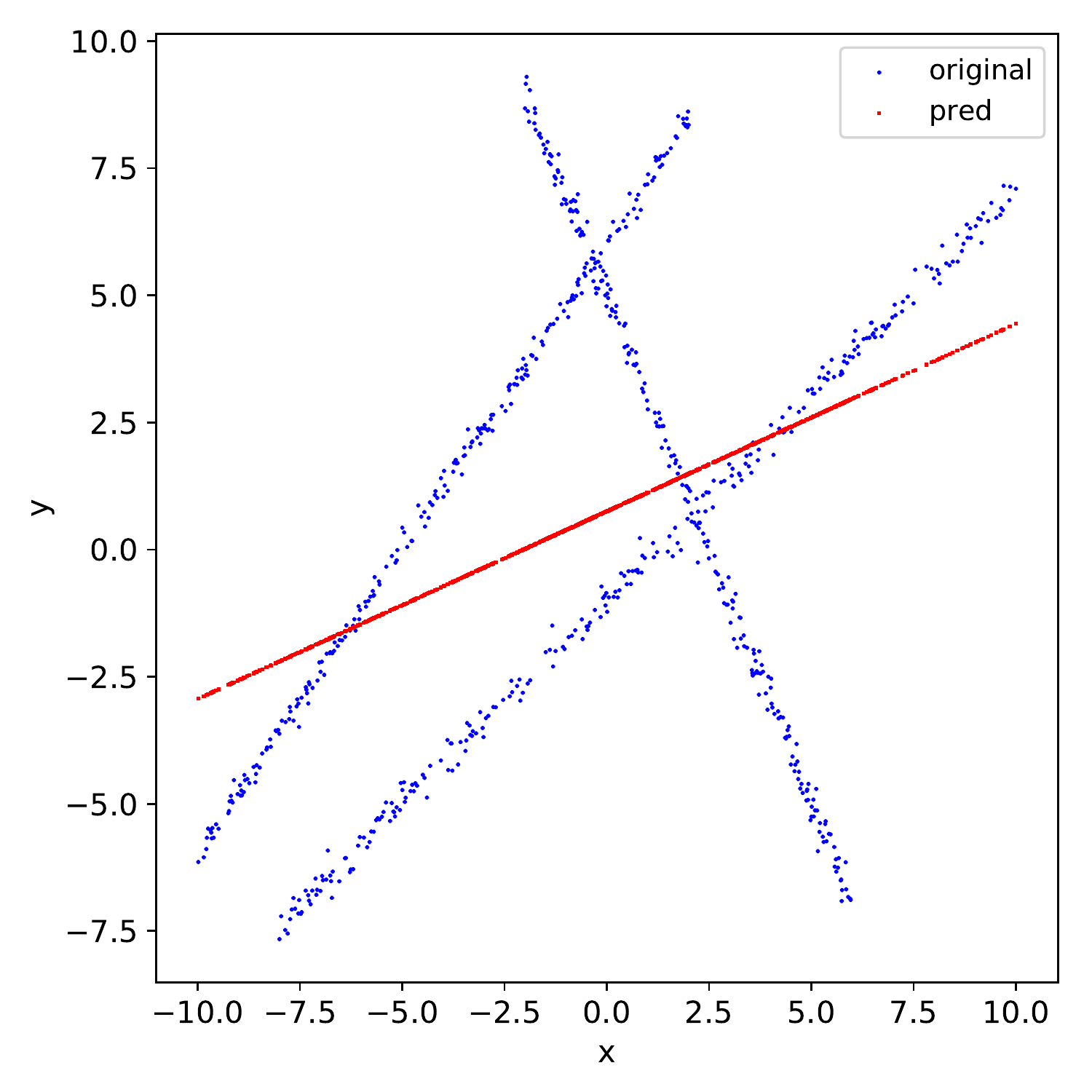}
    \caption{LR} \label{sfig:3lines_linear}
\end{subfigure}
\begin{subfigure}[b]{0.3\linewidth}
    \includegraphics[width=\linewidth]{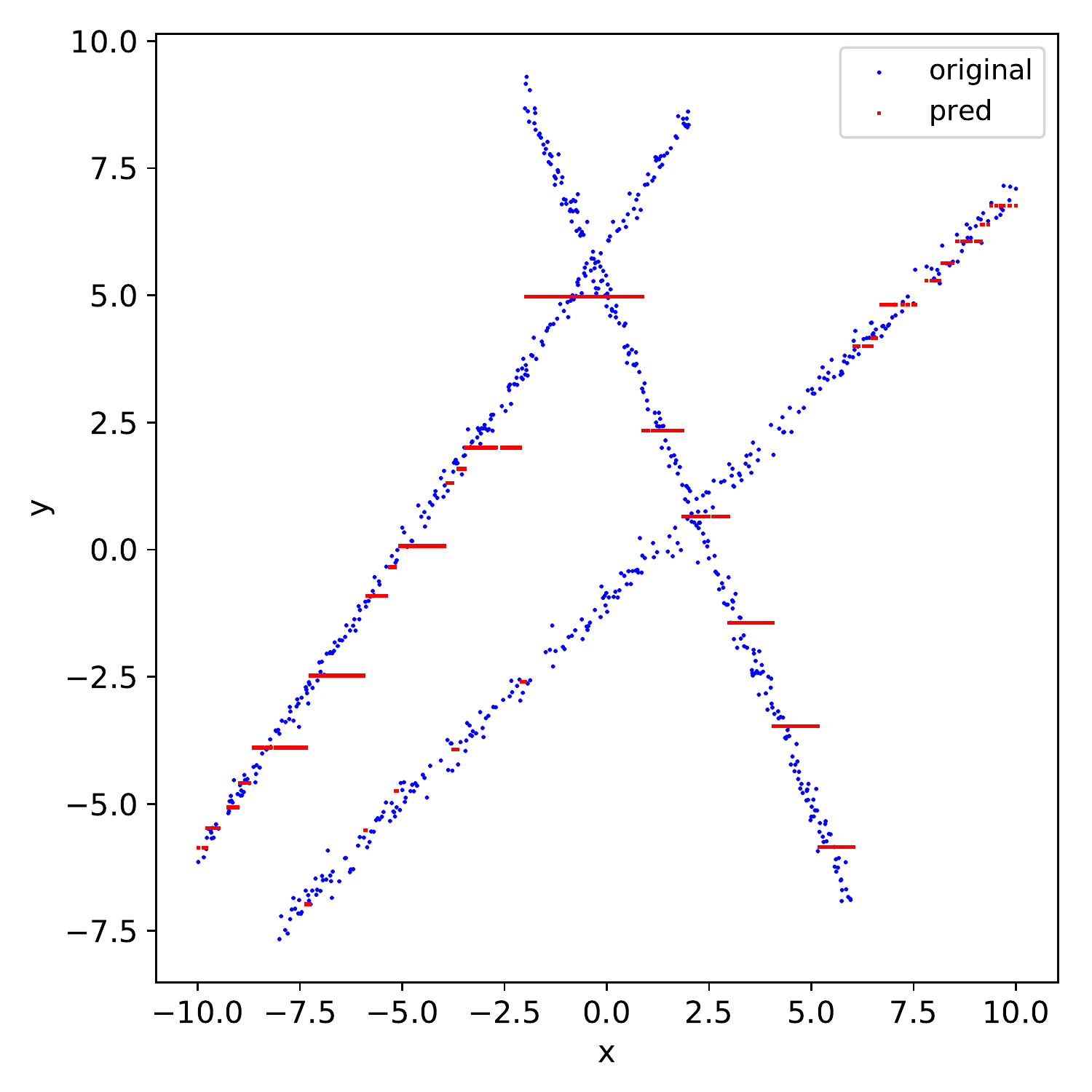}
    \caption{DT} \label{sfig:3lines_dt}
\end{subfigure}    
\begin{subfigure}[b]{0.3\linewidth}
    \includegraphics[width=\linewidth]{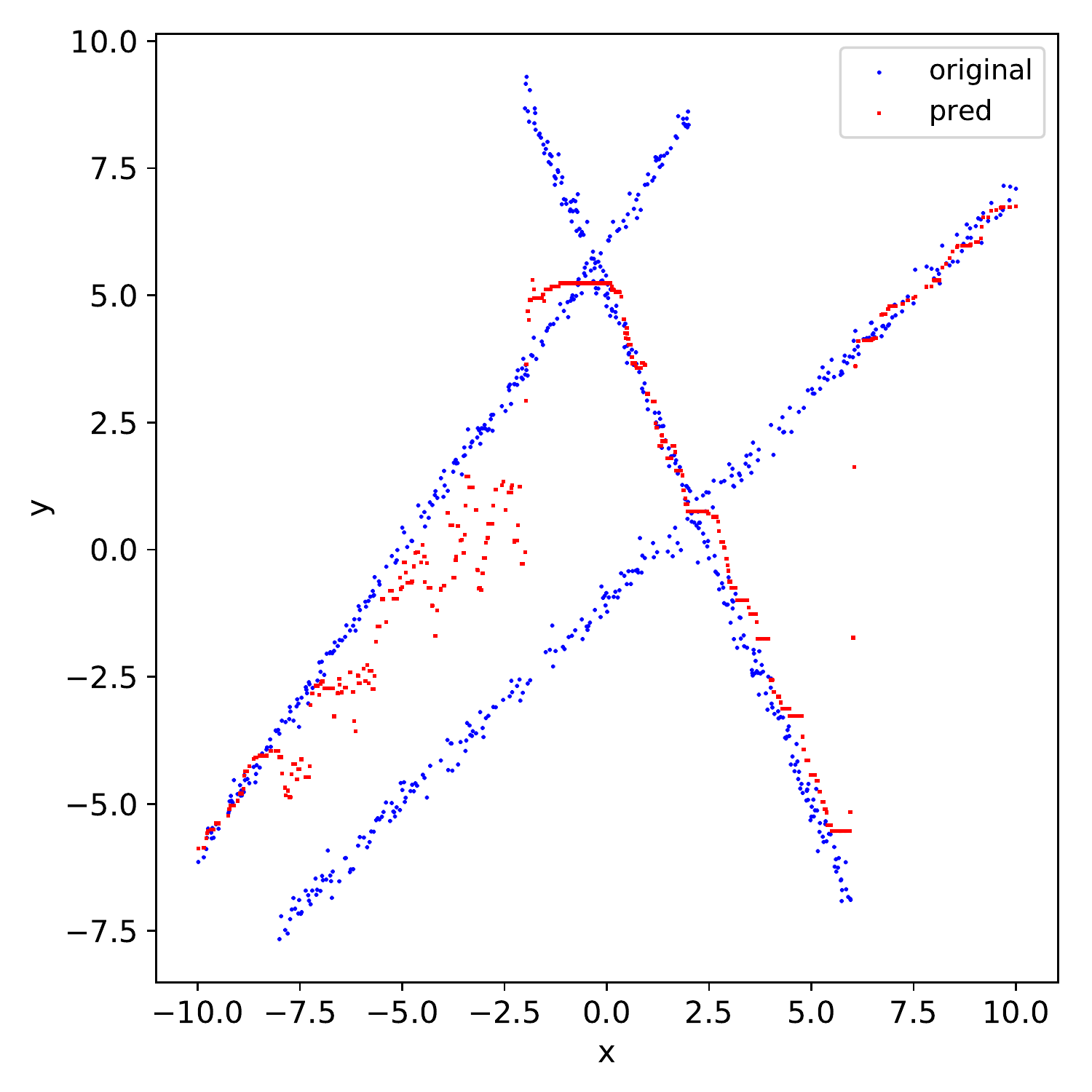}
    \caption{RF} \label{sfig:3lines_rf}
\end{subfigure} 
\hfill
\begin{subfigure}[b]{0.3\linewidth}
    \includegraphics[width=\linewidth]{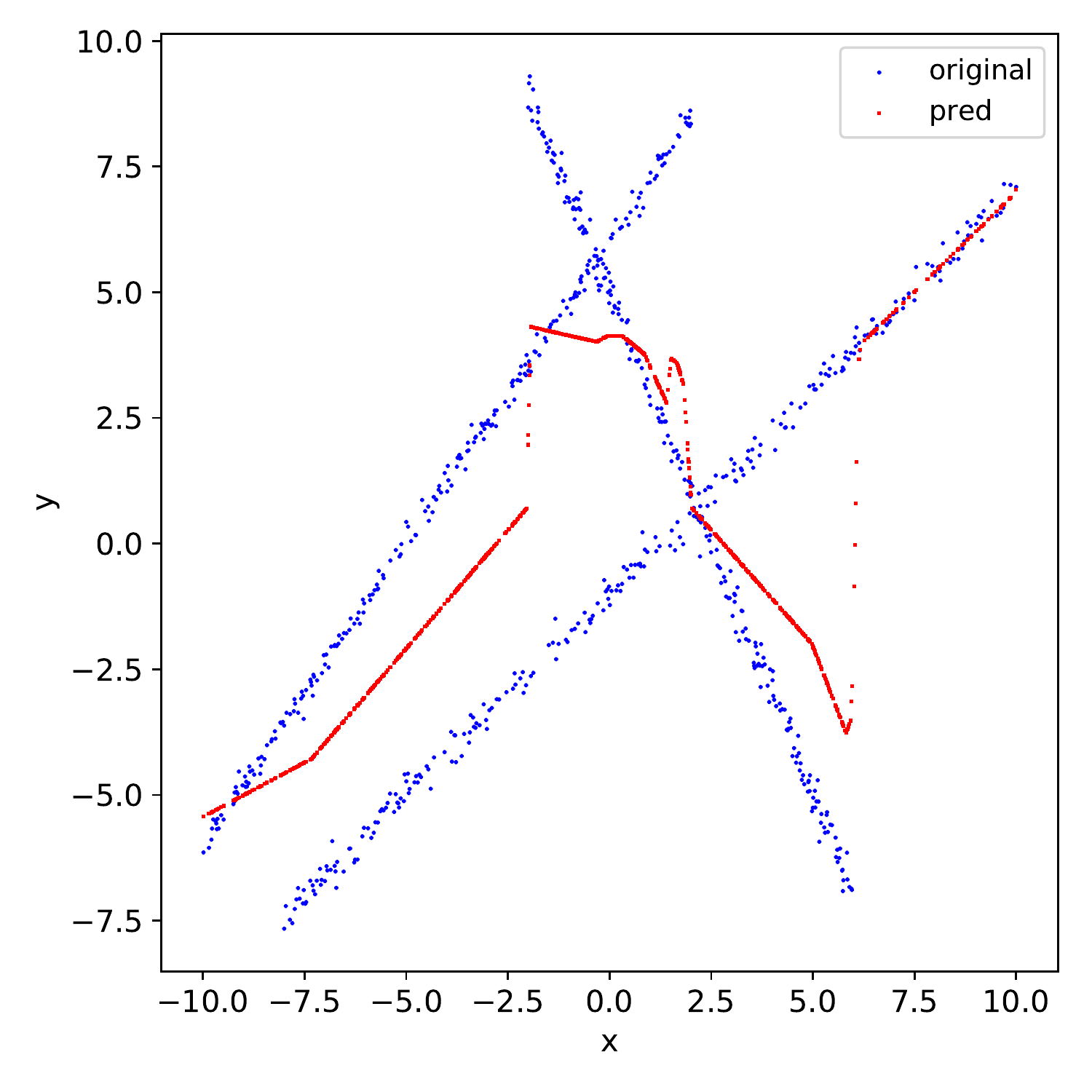}
    \caption{MLP} \label{sfig:3lines_mlp}
\end{subfigure}
\begin{subfigure}[b]{0.3\linewidth}
    \includegraphics[width=\linewidth]{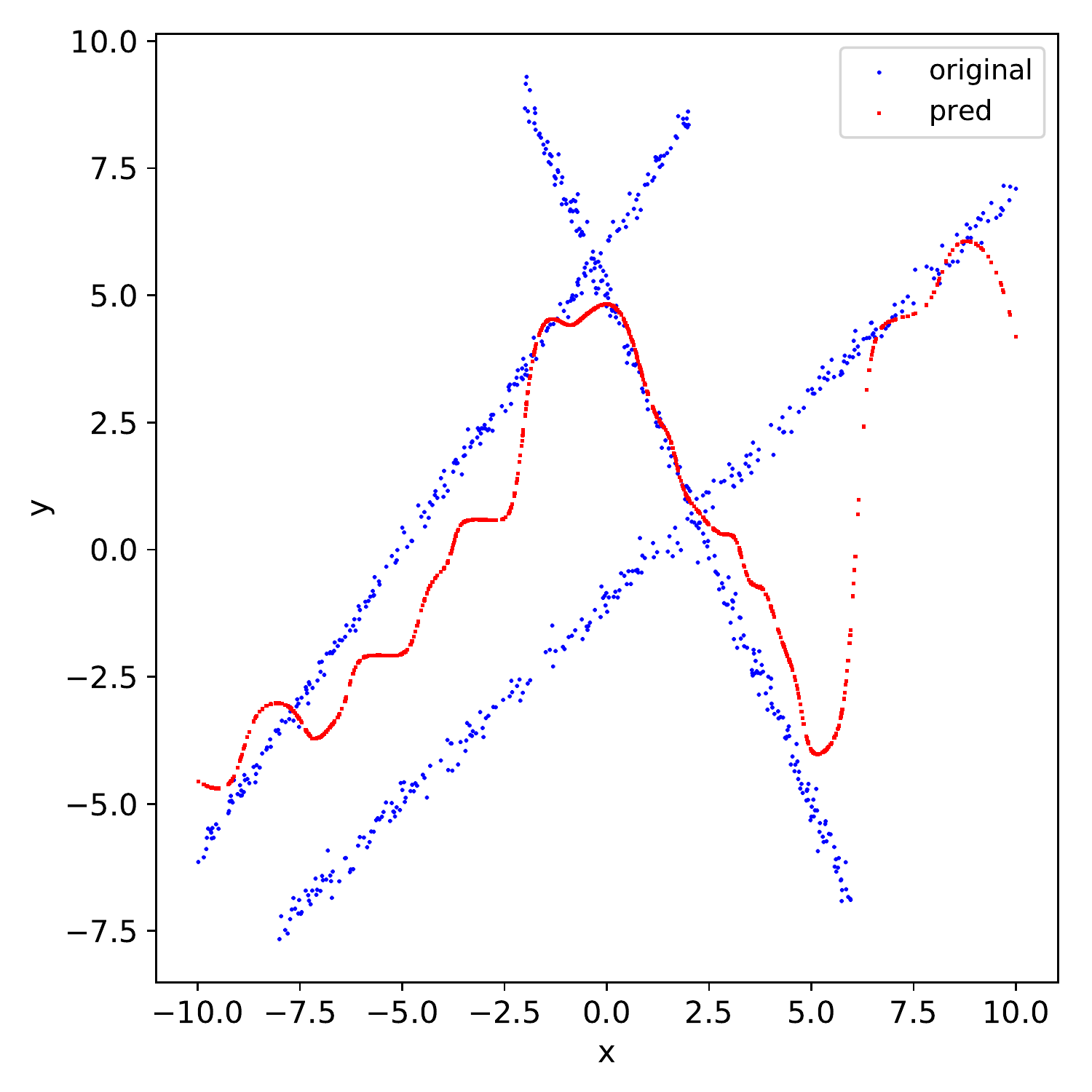}
    \caption{HRME-LR} \label{sfig:3lines_tree_linear}
\end{subfigure}    
\begin{subfigure}[b]{0.3\linewidth}
    \includegraphics[width=\linewidth]{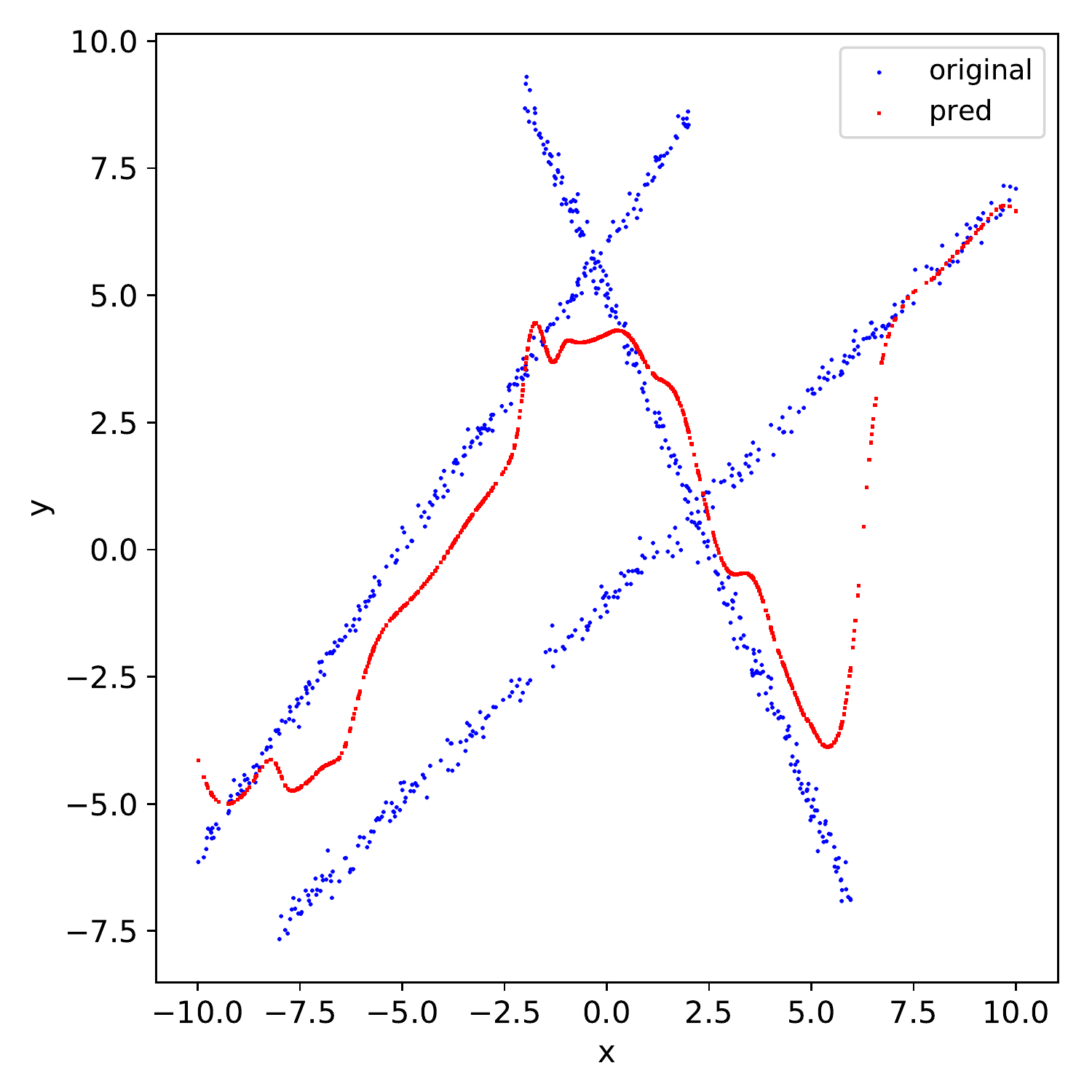}
    \caption{HRME-SVR} \label{sfig:3lines_tree_svr}
\end{subfigure}
\caption{Fitting results on synthetic data with different models: linear regression (LR), decision tree (DT), random forest (RF), multi-layer perceptron (MLP), our HRME with linear regressor (HRME-LR) and SVR regressor (HRME-SVR), respectively.} \label{fig:3lines_plots}
\end{figure}

On the synthetic 3-lines data, Figure~\ref{fig:3lines_plots} shows the fitting results on the test set for our methods and baseline methods. We observe that our HRME models provide a more accurate prediction than the baselines. Specifically, the linear model is just predicting the mean of the three different distributions; the decision tree and random forest provide a better fit than linear regression, but discontinuities and higher variance occur due to the piece-wise linear nature of these two models.
MLP achieves smaller prediction error than DT and RF, but it also shows discontinuities and failure to capture the data modality. 
In comparison, our HRME models provide much smoother fitting with lower bias and variance than the baselines.
Note that even with linear leaf experts, the HRME-LR model is able to capture the nonlinear modality of the data and make regional predictions by soft-switching its experts among the three distributions.
Further, by using non-linear leaf experts, the HRME-SVR model yields smoother predictions than the HRME-LR model with lower bias and variance.
Additionally, we observe that all models here prefer the upper line to the lower line due to the higher noise level in the lower line.



Figure~\ref{sfig:3lines_plots_expert} shows the predictions made by the experts in HRME-SVR model. We see that there are total fourteen experts (indicated by colored curves) being allocated
to different regions of the data. Each expert is confident of making predictions within one data mode, as indicated by higher posterior probabilities (darker colors), and all data modes are successfully captured.
Consequently, if we have prior knowledge of the data distribution, this could be used to select the experts for making the best predictions.
Further, instead of using weighted-average over all experts, we select the top-$1$ expert to make predictions.
Figure~\ref{sfig:3lines_plots_expert_top} shows the corresponding fitting results. We see a much better fit than that in Figure~\ref{sfig:3lines_tree_svr}---in the former all data modes are successfully predicted by our HRME-SVR model.

\begin{figure}[!t]
\centering
{\footnotesize
\scalebox{0.8}{
\begin{forest}
for tree={fit=tight, minimum width=2.5em, l sep=1em, s sep=1.25em, anchor=center}
[
[-1.9, edge label={node[midway, left, red]{2.86}}, circle, draw  
    [-3.09, edge label={node[midway, left, red]{2.85}}, circle, draw
        [-7, edge label={node[midway, left, red]{2.846}}, circle, draw
            [, edge={dashed}]
            [-3.7, edge label={node[midway, right, red]{2.845}}, circle, draw
                [-6.4, edge label={node[midway, left, red]{2.844}}, circle, draw
                    [, edge={dashed}]
                    [-4.3, edge label={node[midway, right, red]{2.843}}, circle, draw
                        [-4.9, edge label={node[midway, left, red]{2.843}}, circle, draw
                            [, edge={dashed}]
                            [, edge={dashed}]
                        ]
                        [, edge={dashed}]
                    ]
                ]
                [, edge={dashed}]
            ]
        ]
        [-2.5, edge label={node[midway, right, red]{2.842}}, circle, draw
            [, edge={dashed}]
            [, edge={dashed}]
        ]
    ]
    [7.7, edge label={node[midway, right, red]{2.827}}, circle, draw
        [5.3, edge label={node[midway, left, red]{2.822}}, circle, draw
            [, edge={dashed}]
            [6.8, edge label={node[midway, right, red]{2.822}}, circle, draw
                [5.6, edge label={node[midway, left, red]{2.821}}, circle, draw
                    [, edge={dashed}]
                    [5.9, edge label={node[midway, right, red]{2.820}}, circle, draw
                        [, edge={dashed}]
                        [, edge={dashed}]
                    ]
                ]
                [, edge={dashed}]
            ]
        ]
        [, edge={dashed}]
    ]
]
]
\end{forest}
}
}
\caption{The HRME tree after training on the synthetic data. The tree is grown recursively in a \emph{depth-first} manner---top to bottom, left to right. Each circle represents a classifier node, and the number within it is the partition threshold $t$. The number on the edge represents the root mean square error if stop growing at that node. Each dashed edge leads to a leaf regressor.} \label{fig:3lines_mse_tree}
\end{figure}
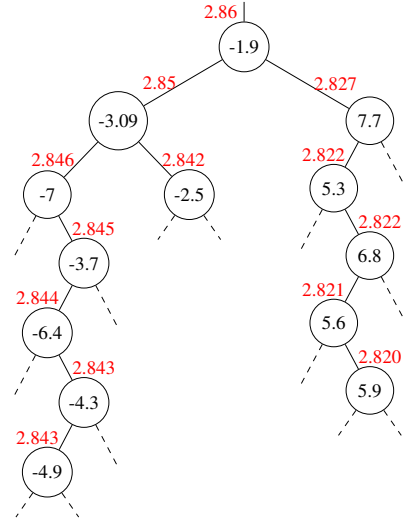

We further show the growth of the HRME tree on the training set.
In Figure~\ref{fig:3lines_mse_tree}, the number in each circle node is the partition threshold $t$. The number besides each circle is the RMSE if growth stops at that node. Note the tree is grown recursively in a depth-first manner (top to bottom, left to right). We observe that the RMSE reduces as the tree grows. This validates our hypothesis that our algorithm can learn the optimal tree structure automatically without pruning afterwards, and the proposed $Q$-value is a good indicator of the global optimality of the tree.
Further, we notice our HRME model also successfully partitions the output space based on separability of data modes by finding the thresholds like $-1.9$, $5.6$, $-6.4$, etc.

Table~\ref{tab:results} shows comprehensive results for all the methods on all the datasets.
We observe an overall improvement of our HRME methods over the baseline methods. Specifically, for large datasets like Engery and Kin40k, our methods outperform all other baselines in terms of both bias (MAE) and variance (RMSE)
even for the HME models with strong Gaussian process experts and the MLP.
For medium-sized datasets like CCPP and Concrete our methods generally outperform other baselines except RF.
But as an ensemble method like RF, our method can also be boosted (now is averaged) to improve performance~\cite{avnimelech1999boosted}.
For small dataset like Housing, our methods do not outperform DT and RF. But at a closer look we find that HRME-LR yields much smaller MAE and RMSE than HRME-SVR and is on a par with DT and RF. This observation indicates the linear nature of data distributions, and hence a nonlinear regression expert would be inappropriate for this dataset. This observation is also confirmed by the poor performance of the nonlinear MLP model.
Further, the data is small ($506$ samples) but has high dimension ($13$), making it difficult to separate the modes by SVM. Instead, other classifiers can be used to improve the performance of our model. 
We also observe that our methods can reduce the variance (low RMSE) on a majority of tasks. This shows that our methods are able to mitigate the problem of high variance of conventional tree models. 
In addition, we see even with simple linear leaf experts, our method can significantly outperform LR, and can compete with nonlinear models like SVR, RF and MLP. This validates our hypothesis that with our data modality-aware routing mechanism simple leaf experts can make good predictions.
At last, The MLP performs poorly in most of the tasks even with fine-tuning. This shows that MLP is not able to capture the complex modality of data distributions.

To this point, comprehensive experiment results show that our HRME methods perform well on a wide range of regression tasks, especially on large, high-dimensional and difficult datasets. Our HRME methods can capture the complex data hierarchy, reduce variance, and make good predictions with simple leaf experts.
We further explore some theoretical properties of our HRME model.

\begin{savenotes}
\begin{table*}[!t]
\caption{Experiment Results}
\label{tab:results}
\vskip 0.5 \baselineskip
\begin{center}
\scalebox{1}{
\begin{sc}
\begin{tabular}{lcccccccccc}
\toprule

Dataset & Metric & \multicolumn{1}{c}{LR} & \multicolumn{1}{c}{SVR} & \multicolumn{1}{c}{DT} & \multicolumn{1}{c}{RF} & \multicolumn{1}{c}{HME} & \multicolumn{1}{c}{MLP} & \multicolumn{2}{c}{{\bf HRME}} \\
& & & & & & & & \multicolumn{1}{c}{LR} & \multicolumn{1}{c}{SVR} \\

\midrule
3-lines & \begin{tabular}{@{}c@{}} MAE \\ RMSE \end{tabular} 
& \begin{tabular}{@{}c@{}} 3.352 \\ 4.104 \end{tabular} 
& \begin{tabular}{@{}c@{}} 2.006 \\ 3.173 \end{tabular} 
& \begin{tabular}{@{}c@{}} 2.224 \\ 3.291 \end{tabular} 
& \begin{tabular}{@{}c@{}} 2.131 \\ 3.072 \end{tabular} 
& \begin{tabular}{@{}c@{}}  ---  \\  ---  \end{tabular} 
& \begin{tabular}{@{}c@{}} 1.960 \\ 2.795 \end{tabular} 
& \begin{tabular}{@{}c@{}} 2.337 \\ 2.885 \end{tabular} 
& \begin{tabular}{@{}c@{}} 2.250 \\ 2.859 \end{tabular}\\

\midrule
Housing & \begin{tabular}{@{}c@{}} MAE \\ RMSE \end{tabular} 
& \begin{tabular}{@{}c@{}} 3.651 \\ 4.911 \end{tabular} 
& \begin{tabular}{@{}c@{}} 3.498 \\ 5.126 \end{tabular} 
& \begin{tabular}{@{}c@{}} 2.537 \\ 3.665 \end{tabular} 
& \begin{tabular}{@{}c@{}} 2.103 \\ 3.043 \end{tabular} 
& \begin{tabular}{@{}c@{}} 4.170~\footnote{using Gaussian experts; results taken from~\citet{ferrari2011choices}} \\ 5.610~\footnote{using Gaussian experts; results taken from~\citet{ferrari2011choices}} \end{tabular} 
& \begin{tabular}{@{}c@{}} 6.711 \\ 8.535 \end{tabular} 
& \begin{tabular}{@{}c@{}} 2.682 \\ 3.857 \end{tabular} 
& \begin{tabular}{@{}c@{}} 3.266 \\ 4.376 \end{tabular}\\

\midrule
Concrete & \begin{tabular}{@{}c@{}} MAE \\ RMSE \end{tabular} 
& \begin{tabular}{@{}c@{}} 8.088 \\ 10.204 \end{tabular} 
& \begin{tabular}{@{}c@{}} 8.013 \\ 10.772 \end{tabular} 
& \begin{tabular}{@{}c@{}} 4.919 \\  8.000 \end{tabular} 
& \begin{tabular}{@{}c@{}} 3.436 \\  4.806 \end{tabular} 
& \begin{tabular}{@{}c@{}}  ---  \\  6.250~\footnote{using Gaussian process experts; results taken from~\citet{trapp2018learning}} \end{tabular} 
& \begin{tabular}{@{}c@{}} 5.394 \\  6.594 \end{tabular} 
& \begin{tabular}{@{}c@{}} 4.121 \\  5.664 \end{tabular} 
& \begin{tabular}{@{}c@{}} 4.020 \\  5.609 \end{tabular}\\

\midrule
CCPP & \begin{tabular}{@{}c@{}} MAE \\ RMSE \end{tabular} 
& \begin{tabular}{@{}c@{}} 3.601 \\ 4.578 \end{tabular} 
& \begin{tabular}{@{}c@{}} 2.746 \\ 3.856 \end{tabular} 
& \begin{tabular}{@{}c@{}} 2.941 \\ 4.151 \end{tabular} 
& \begin{tabular}{@{}c@{}} 2.383 \\ 3.409 \end{tabular} 
& \begin{tabular}{@{}c@{}}  ---  \\ 4.100~\footnote{using Gaussian process experts; results taken from~\citet{trapp2018learning}} \end{tabular} 
& \begin{tabular}{@{}c@{}} 4.013 \\ 5.078 \end{tabular} 
& \begin{tabular}{@{}c@{}} 2.965 \\ 3.951 \end{tabular} 
& \begin{tabular}{@{}c@{}} 2.712 \\ 3.805 \end{tabular}\\

\midrule
Energy & \begin{tabular}{@{}c@{}} MAE \\ RMSE \end{tabular} 
& \begin{tabular}{@{}c@{}} 52.075 \\  93.564 \end{tabular} 
& \begin{tabular}{@{}c@{}} 43.141 \\ 101.267 \end{tabular} 
& \begin{tabular}{@{}c@{}} 43.996 \\  99.654 \end{tabular} 
& \begin{tabular}{@{}c@{}} 52.002 \\  95.558 \end{tabular} 
& \begin{tabular}{@{}c@{}}   ---  \\   ---  \end{tabular} 
& \begin{tabular}{@{}c@{}} 40.521 \\ 88.191 \end{tabular} 
& \begin{tabular}{@{}c@{}} 42.121 \\ 89.203 \end{tabular} 
& \begin{tabular}{@{}c@{}} 40.009 \\ 87.022 \end{tabular}\\

\midrule
Kin40k & \begin{tabular}{@{}c@{}} MAE \\ RMSE \end{tabular} 
& \begin{tabular}{@{}c@{}} 0.806 \\ 0.996 \end{tabular} 
& \begin{tabular}{@{}c@{}} 0.092 \\ 0.161 \end{tabular} 
& \begin{tabular}{@{}c@{}} 0.592 \\ 0.773 \end{tabular} 
& \begin{tabular}{@{}c@{}} 0.433 \\ 0.548 \end{tabular} 
& \begin{tabular}{@{}c@{}}  ---  \\ 0.230~\footnote{using Gaussian process experts; results taken from~\citet{nguyen2014fast}} \end{tabular} 
& \begin{tabular}{@{}c@{}} 0.237 \\ 0.312 \end{tabular} 
& \begin{tabular}{@{}c@{}} 0.150 \\ 0.212 \end{tabular} 
& \begin{tabular}{@{}c@{}} 0.071 \\ 0.114 \end{tabular}\\

\bottomrule
\end{tabular}
\end{sc}
}
\end{center}
\end{table*}
\end{savenotes}

\subsection{Convergence and Complexity Analysis}
\emph{Convergence:}
Let $n$, $d$, $k$ be the number of training samples, the dimension of each sample and the number of experts, respectively. \citet{zeevi1998error} prove that with large samples, the ME models can uniformly approximate Sobolev class functions of order $r$ in the $L_p$ norm at a rate of at least $\gO(Ck^{-r/d})$ with constant $C$. This upper-bounds the approximation error of general ME family.
Further, \citet{jiang1999approximation} prove that the HME mean functions can approximate the true mean function at a rate of $\gO(k^{-2/d})$ in the $L_p$ norm.
\citet{jiang1999hierarchical} also show that the HME probability density functions can approximate the data density at a rate of $\gO(k^{-4/d})$ in KL divergence.
For our HRME model, since the general assumptions of these results hold, the uniform convergence also holds.

\emph{Complexity:} 
The complexity of EM based algorithms for HME models mainly lies in the M-step, where the re-estimation of parameters involves solving a system of equations using the Newton (or Newton-like) update.
In the HME models, a Newton iteration cost is $\gO(n^3)$.
In our case, the complexity of M-step is in solving the SVM.
Specifically, for standard SVM solver with primal-dual interior point method, the complexity is in the Newton update and evaluation of the kernel, and hence the iteration cost is $\gO(n^3 + n^2d)$.
As a result, to attain $\epsilon$-error we need $\gO(\epsilon^{-d/2})$ experts.
For the HME models, we can assume uniform data partition among experts, and the total cost is $\gO(n^3\epsilon^d)$.
For our HRME model, the data for each node decreases with depth, and we can take the average among nodes. The resultant total cost is $\gO(n^3\epsilon^d + dn^2\epsilon^{d/2})$.
Although the total complexity increases for our algorithm, however, the computation can be accelerated using dynamic programming at the price of storage cost. Moreover, the computation at each node can be done in parallel.

\emph{Consistency:} 
\citet{zeevi1998error} prove that under regularity conditions, least-squares estimators for the ME models are consistent.
Further, \citet{jiang2000asymptotic} show that maximum likelihood estimators are consistent and asymptotically normal.
Therefore, our HRME model also produces consistent estimators.

\emph{Identifiability:}
\citet{jiang1999identifiability} prove that the ME models are identifiable under regularity conditions that the experts are ordered and the model parameters are carefully initialized.

In the future work, we would like to provide more rigorous study on the theoretically behaviors of our HRME model.

\section{Conclusions}
In this paper, we propose a hierarchical routing mixture of experts (HRME) model to address the difficulty of data partitioning and expert assigning in conventional regression models.
By utilizing non-leaf classifier experts, our model is able to capture the natural data hierarchy and route the data to simple regressors for effective predictions.
Furthermore, we develop a probabilistic framework for the HRME model, and propose a recursive Expectation-Maximization (EM) based algorithm to optimize both the tree structure as well as the expert models.
Comprehensive experiment results validate the effectiveness and some nice properties of our model.







\bibliography{icml2019_HRME}
\bibliographystyle{icml2019}

\end{document}